\documentclass[11pt,a4paper]{article}
\usepackage[hyperref]{emnlp2020}
\usepackage{times}
\usepackage{latexsym}
\usepackage{graphicx}
\usepackage{subfigure}
\usepackage{multirow}
\usepackage{amsmath}
\usepackage{algorithm}
\usepackage{algorithmic}

\usepackage{microtype}

\aclfinalcopy 
\def\aclpaperid{16} 

\title{Meta Fine-Tuning Neural Language Models for Multi-Domain Text Mining}

\author{Chengyu Wang$^{1}$, Minghui Qiu$^1$\thanks{\ \ Corresponding author.}, Jun Huang$^1$, Xiaofeng He$^{2}$\\
 $^1$ Alibaba Group $^2$ East China Normal University\\
 {\tt \{chengyu.wcy, minghui.qmh, huangjun.hj\}@alibaba-inc.com} \\
 {\tt hexf@cs.ecnu.edu.cn}}

\date{}

\begin{document}
\maketitle
\begin{abstract}
Pre-trained neural language models bring significant improvement for various NLP tasks, by fine-tuning the models on task-specific training sets. During fine-tuning, the parameters are initialized from pre-trained models directly, which ignores how the learning process of similar NLP tasks in different domains is correlated and mutually reinforced. In this paper, we propose an effective learning procedure named \emph{Meta Fine-Tuning} (MFT), serving as a meta-learner to solve a group of similar NLP tasks for neural language models. Instead of simply multi-task training over all the datasets, MFT only learns from typical instances of various domains to acquire highly transferable knowledge. It further encourages the language model to encode domain-invariant representations by optimizing a series of novel domain corruption loss functions. After MFT,  the model can be fine-tuned for each domain with better parameter initialization and higher generalization ability. We implement MFT upon BERT to solve several multi-domain text mining tasks. Experimental results confirm the effectiveness of MFT and its usefulness for few-shot learning.
\end{abstract}

\section{Introduction}

Recent years has witnessed a boom in pre-trained neural language models. Notable works include ELMo~\cite{DBLP:conf/naacl/PetersNIGCLZ18}, BERT~\cite{DBLP:conf/naacl/DevlinCLT19}, Transformer-XL~\cite{DBLP:conf/acl/DaiYYCLS19}, ALBERT~\cite{DBLP:journals/corr/abs-1909-11942}, StructBERT~\cite{DBLP:journals/corr/abs-1908-04577} and many others. These models revolutionize the learning paradigms of various NLP tasks. After pre-training, only a few \emph{fine-tuning} epochs are required to train models for these tasks. 

The ``secrets'' behind this phenomenon owe to the models' strong representation learning power to encode the semantics and linguistic knowledge from massive text corpora~\cite{DBLP:conf/acl/JawaharSS19,DBLP:conf/emnlp/KovalevaRRR19,DBLP:conf/naacl/Liu0BPS19,DBLP:conf/acl/TenneyDP19}. By simple fine-tuning, models can transfer the universal Natural Language Understanding (NLU) abilities to specific tasks~\cite{DBLP:conf/iclr/WangSMHLB19}. However, state-of-the art language models mostly utilize \emph{self-supervised} tasks during pre-training (for instance, masked language modeling and next sentence prediction in BERT~\cite{DBLP:conf/naacl/DevlinCLT19}). This unavoidably creates a~\emph{learning gap} between pre-training and fine-tuning. Besides, for a group of similar tasks, conventional practices require the parameters of all task-specific models to be initialized from the same pre-trained language model, ignoring how the learning process in different domains is correlated and mutually reinforced. 

A basic solution is fine-tuning models by multi-task learning. Unfortunately, multi-task fine-tuning of BERT does not necessarily yield better performance across all the tasks~\cite{DBLP:conf/cncl/SunQXH19}. A probable cause is that learning too much from other tasks may force the model to acquire non-transferable knowledge, which harms the overall performance. A similar finding is presented in~\citet{DBLP:conf/eacl/SogaardB17,DBLP:journals/corr/abs-1806-08730} on multi-task training of neural networks. Additionally, language models such as BERT do not have the ``shared-private'' architecture~\cite{DBLP:conf/acl/LiuQH17} to enable effective learning of domain-specific and domain-invariant features. 
Other approaches modify the structures of language models to accommodate multi-task learning and mostly focus on specific applications, without providing a unified solution for all the tasks~\cite{DBLP:conf/icml/Stickland019,DBLP:journals/corr/abs-1910-14296,DBLP:journals/corr/abs-2002-02450}.

\begin{figure}
\centering
\includegraphics[width=.475\textwidth]{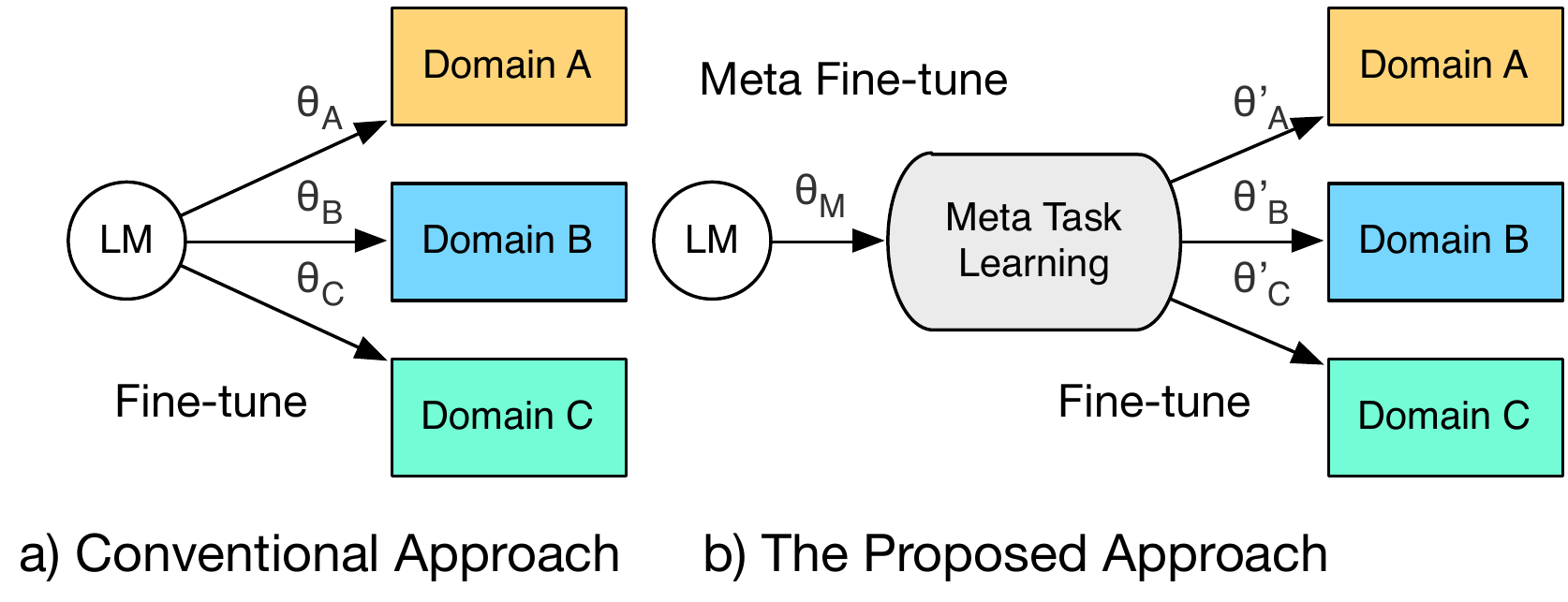}
\caption{Comparison between fine-tuning and MFT. ``LM'' refers to pre-trained language models.}
\label{fig:compare}
\end{figure}

A recent study~\cite{DBLP:conf/icml/FinnAL17} reveals that meta-learning achieves better parameter initialization for a group of tasks, which improves the models' generalization abilities in different domains and makes them easier to fine-tune. 
As pre-trained language models have general NLU abilities, they should also have the ability to learn solving a group of similar NLP tasks.
In this work, we propose a separate learning procedure, inserted between pre-training and fine-tuning, named \emph{Meta Fine-Tuning} (MFT). This work is one of the early attempts for improving fine-tuning of neural language models by meta-learning.
Take the review analysis task as an example. MFT only targets at learning the polarity of reviews (positive or negative) in general, ignoring features of specific aspects or domains. After that, the learned model can be adapted to any domains by fine-tuning. The comparison between fine-tuning and MFT is shown in Figure~\ref{fig:compare}. Specifically, MFT first learns the embeddings of \emph{class prototypes} from multi-domain training sets, and assigns~\emph{typicality scores} to individuals, indicating the transferability of each instance. Apart from minimizing the multi-task classification loss over typical instances, MFT further encourages the language model to learn \emph{domain-invariant} representations by jointly optimizing a series of novel \emph{domain corruption loss} functions.

For evaluation, we implement the MFT strategy upon BERT~\cite{DBLP:conf/naacl/DevlinCLT19} for three multi-domain text mining tasks: i) natural language inference~\cite{DBLP:conf/naacl/WilliamsNB18} (sentence-pair classification), ii) review analysis~\cite{DBLP:conf/acl/BlitzerDP07} (sentence classification) and iii) domain taxonomy construction~\cite{DBLP:conf/emnlp/LuuTHN16} (word-pair classification). Experimental results show that the effectiveness and superiority of MFT. We also show that MFT is highly useful for multi-domain text mining in the few-shot learning setting.
\footnote{Although we focus on MFT for BERT only, MFT is general and can be applied to other language models easily.}

\begin{figure*}
\centering
\includegraphics[width=.975\textwidth]{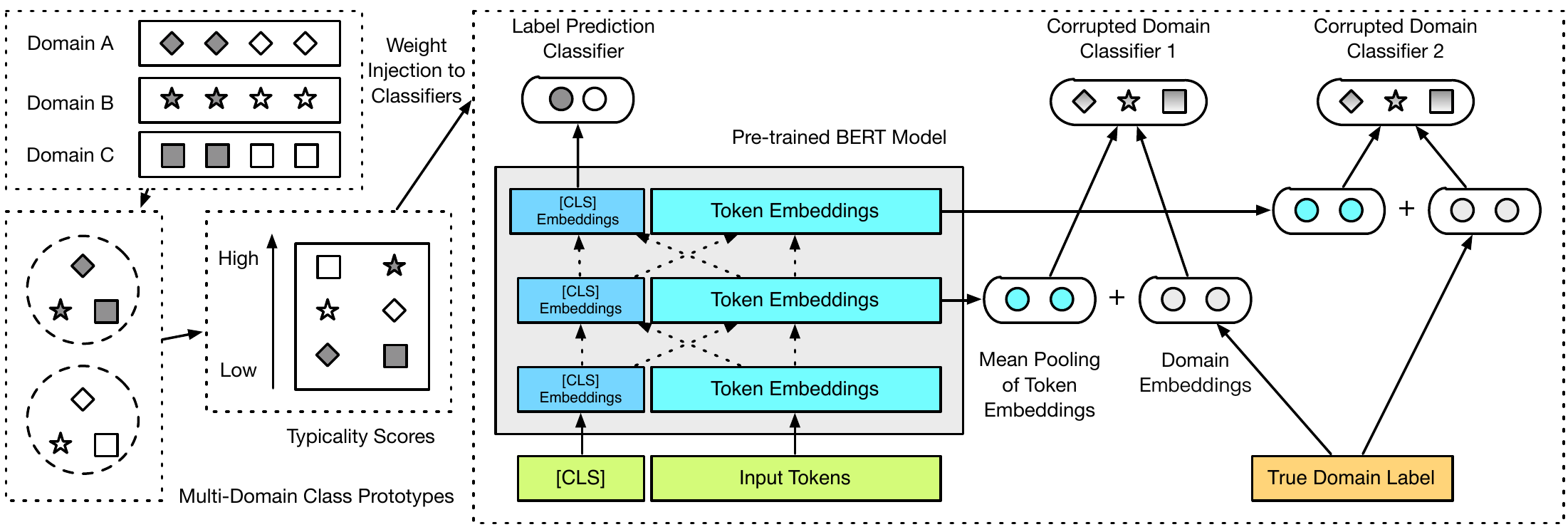}
\caption{The neural architecture of MFT for BERT~\cite{DBLP:conf/naacl/DevlinCLT19}. Due to space limitation, we only show two corrupted domain classifiers and three layers of transformer encoders, with others omitted.}
\label{fig:model}
\end{figure*}

\section{Related Work}
We overview recent advances on pre-trained language models, transfer learning and meta-learning.

\subsection{Pre-trained Language Models}

Pre-trained language models have gained much attention from the NLP community recently~\cite{DBLP:journals/corr/abs-2003-08271}. Among these models, ELMo~\cite{DBLP:conf/naacl/PetersNIGCLZ18} learns context-sensitive embeddings for each token form both left-to-right and right-to-left directions. BERT~\cite{DBLP:conf/naacl/DevlinCLT19} is usually regarded as the most representative work, employing transformer encoders to learn language representations. The pre-training technique of BERT is improved in~\citet{DBLP:journals/corr/abs-1907-11692}. Follow-up works employ transformer-based architectures, including Transformer-XL~\cite{DBLP:conf/acl/DaiYYCLS19}, XLNet~\cite{DBLP:conf/nips/YangDYCSL19}, ALBERT~\cite{DBLP:journals/corr/abs-1909-11942}, StructBERT~\cite{DBLP:journals/corr/abs-1908-04577} and many more. They change the unsupervised learning objectives of BERT in pre-training. 
MT-DNN~\cite{DBLP:conf/acl/LiuHCG19} is the representative of another type of pre-trained language models, which employs supervised learning objectives across tasks to learn representations.

After language models are pre-trained, they can be fine-tuned for a variety of NLP tasks. The techniques of fine-tuning BERT are summarized in~\citet{DBLP:conf/cncl/SunQXH19}.~\citet{DBLP:conf/emnlp/CuiLCZ19} improve BERT's fine-tuning by sparse self-attention.~\citet{DBLP:conf/emnlp/AraseT19} introduce the concept of ``transfer fine-tuning'', which injects phrasal paraphrase relations into BERT. Compared to previous methods,  fine-tuning for multi-domain learning has not been sufficiently studied.

\subsection{Transfer Learning}

Transfer learning aims to transfer the resources or models from one domain (the source domain) to another (the target domain), in order to improve the model performance of the target domain. Due to space limitation, we refer readers to the surveys~\cite{DBLP:journals/tkde/PanY10,DBLP:journals/kbs/LuBHZXZ15,DBLP:journals/corr/abs-1911-02685} for an overview. For NLP applications,  the ``shared-private'' architecture~\cite{DBLP:conf/acl/LiuQH17} is highly popular, which include sub-networks for learning domain-specific representations and a shared sub-network for knowledge transfer and domain-invariant representation learning. Recently, adversarial training has been frequently applied~\cite{DBLP:conf/aaai/ShenQZY18,DBLP:conf/emnlp/HuWZGCS19,DBLP:conf/emnlp/Cao000L18,DBLP:conf/emnlp/LiLWBZY19,DBLP:conf/acl/ZhouZJZFGK19}, where the domain adversarial classifiers are trained to help the models to learn domain-invariant features. Multi-domain learning is a special case of transfer learning whose goal is to transfer knowledge across multiple domains for mutual training reinforcement~\cite{DBLP:conf/aaai/PeiCLW18,DBLP:conf/acl/LiBC19,DBLP:conf/ijcai/Cai019}. Our work also addresses multi-domain learning, but solves the problem from a meta-learning aspect.

\subsection{Meta-learning}

Compared to transfer learning, meta-learning is a slightly different learning paradigm. Its goal is to train meta-learners that can adapt to a variety of different tasks with little training data~\cite{DBLP:journals/corr/abs-1810-03548}, mostly applied to few-shot learning, which is typically formulated as a K-way N-shot learning problem. In NLP, existing meta-learning models mostly focus on training meta-learners for single applications, such as link prediction~\cite{DBLP:conf/emnlp/ChenZZCC19}, dialog systems~\cite{DBLP:conf/acl/MadottoLWF19}, lexical relation classification~\cite{DBLP:journals/corr/abs-2002-10903} and semantic parsing~\cite{DBLP:conf/acl/GuoTDZY19}.~\citet{DBLP:conf/emnlp/DouYA19} leverage meta-learning for low-resource NLU. 

Compared with traditional meta-learning research, the task of our work is not K-way N-shot. Instead, we aim to employ meta-learning to train a better ``meta-learner'' which captures transferable knowledge across domains. 
In this sense, our work can be also viewed as a transfer learning algorithm which employs meta-learning for better knowledge transfer and fast domain adaptation.

\section{MFT: The Proposed Framework}

In this section, we start with some basic notations and an overview of MFT. After that, we describe the algorithmic techniques of MFT in detail.

\subsection{Overview}

Denote $\mathcal{D}^{k}=\{x_i^k, y_i^k\vert i\in[1,N^k]\}$ as the training set of the $k$th domain, where $x_i^k$ and $y_i^k$ are the input text and the class label of the $i$th sample, respectively\footnote{Note that $x_i^k$ can be either a single sentence, a sentence pair, or any other possible input texts of the target NLP task.}. $N^k$ is the number of total samples in $\mathcal{D}^{k}$. 
The goal of MFT is to train a meta-learner initialized from a pre-trained language model, based on the training sets of $K$ domains: $\mathcal{D}=\bigcup_{k=1}^{K}\mathcal{D}^{k}$. The meta-learner provides better parameter initializations, such that it can be easily adapted to each of the $K$ domains by fine-tuning the meta-learner over the training set of the $k$th domain separately.

Due to the large parameter space of neural language models, it is computationally challenging to search for the optimal values of the meta-learner's parameters. As discussed earlier, building a trivial multi-task learner over $\mathcal{D}$ does not guarantee satisfactory results either~\cite{DBLP:conf/cncl/SunQXH19}. Here, we set up two design principles for MFT:~\emph{Learning from Typicality} and~\emph{Learning Domain-invariant Representations}, introduced as follows:

\paragraph{Learning from Typicality}
To make the meta-learner easier to be fine-tuned to any domains, the encoded knowledge should be highly general and transferable, not biased towards specific domains.
Hence, only~\emph{typical} instances w.r.t. all the domains should be the priority learning targets. We first generate~\emph{multi-domain class prototypes} from $\mathcal{D}$ to summarize the semantics of training data. Based on the prototypes, we compute~\emph{typicality scores} for all training instances, treated as weights for MFT.

\paragraph{Learning Domain-invariant Representations} 
A good meta-learner should be adapted to any domains quickly. Since BERT has strong representation power, this naturally motivates us to learn \emph{domain-invariant representations} are vital for fast domain adaptation~\cite{DBLP:conf/aaai/ShenQZY18}. In MFT, besides minimizing the classification loss, we jointly minimize new learning objectives to force the language model to have domain-invariant encoders.

Based on the two general principles, we design the neural architecture of MFT, with the example for BERT~\cite{DBLP:conf/naacl/DevlinCLT19} shown in Figure~\ref{fig:model}. The technical details are introduced subsequently.

\subsection{Learning from Typicality}

Denote $\mathcal{M}$ as the class label set of all $K$ domains. $\mathcal{D}_m^{k}$ is the collection of input texts in $\mathcal{D}^{k}$ that have class label $m\in\mathcal{M}$, i.e.,~$\mathcal{D}_m^{k}=\{x_i^k\vert(x_i^k, y_i^k)\in\mathcal{D}^{k}, y_i^k=m\}$. As class prototypes can summarize the key characteristics of the corresponding data~\cite{DBLP:journals/corr/abs-2001-00745}, we treat the class prototype $\mathbf{c}_m^k$ as the averaged embeddings of all the input texts in $\mathcal{D}_m^{k}$. Formally, we have $\mathbf{c}_m^k=\frac{1}{\vert \mathcal{D}_m^{k}\vert}\sum_{x_i^k\in\mathcal{D}_m^{k}}\mathcal{E}(x_i^k)$ where $\mathcal{E}(\cdot)$ maps $x_i^k$ to its $d$-dimensional embeddings. As for BERT~\cite{DBLP:conf/naacl/DevlinCLT19}, we utilize the mean pooling of representations of $x_i^k$ from the last transformer encoder as $\mathcal{E}(x_i^k)$.

Ideally, we regard a training instance $(x_i^k, y_i^k)$ to be~\emph{typical} if it is semantically close to its class prototype $\mathbf{c}_m^k$, and also is not too far away from class prototypes generated from other domains for high transferability. Therefore, the~\emph{typicality score}~$t_i^k$ of $(x_i^k, y_i^k)$ can be defined as follows:\footnote{Here, we assume that  the training instance $(x_i^k, y_i^k)$ has the class label $m\in\mathcal{M}$. Because each instance is associated with only one typicality score, for simplicity, we denote it as $t_i^k$, instead of $t_{i,m}^k$.}
\begin{equation*}
\begin{split}
t_i^k=&\alpha\cos(\mathcal{E}(x_i^k), \mathbf{c}_m^k)\\
&+\frac{1-\alpha}{K-1}\cdot\sum_{\tilde{k}=1}^{K}\mathbf{1}_{(\tilde{k}\neq k)}\cos(\mathcal{E}(x_i^k), \mathbf{c}_m^{\tilde{k}}),
\end{split}
\end{equation*}
where $\alpha$ is the pre-defined balancing factor ($0<\alpha<1$), $\cos(\cdot,\cdot)$ is the cosine similarity function and $\mathbf{1}_{(\cdot)}$ is the indicator function that returns 1 if the input Boolean function is true and 0 otherwise.

As one prototype may not be insufficient to represent the complicated semantics of the class~\cite{DBLP:conf/acl/CaoHJCL17}, we can also generate multiple prototypes by clustering, with the $j$th prototype to be $\mathbf{c}_{m_j}^{k}$.  Here, $t_i^k$ is extended by the following formula:
\begin{equation*}
\begin{split}
&t_i^k=\alpha\frac{\sum_{n\in\mathcal{M}}\beta_n\cos(\mathcal{E}(x_i^k), \mathbf{c}_{n}^k)}{\sum_{n\in\mathcal{M}}\beta_{n}}\\
&+\frac{1-\alpha}{K-1}\cdot\sum_{\tilde{k}=1}^{K}\frac{\mathbf{1}_{(\tilde{k}\neq k)}\sum_{n\in\mathcal{M}} \beta_n\cos(\mathcal{E}(x_i^k), \mathbf{c}_{n}^{\tilde{k}})}{\sum_{n\in\mathcal{M}}\beta_{n}},
\end{split}
\end{equation*}
where $\beta_n>0$ is the cluster membership of $x_i^k$ w.r.t. each class label $n\in\mathcal{M}$. 

After typicality scores are computed, we discuss how to set up the learning objectives for MFT.
The first loss is the multi-task~\emph{typicality-sensitive label classification loss}~$\mathcal{L}_{TLC}$. It penalizes the text classifier for predicting the labels of typical instances of all $K$ domains incorrectly, which is defined as:\footnote{For clarity, we omit all the regularization terms in objective functions throughout this paper.}
\begin{equation*}
\begin{split}
\mathcal{L}_{TLC}= & -\frac{1}{\vert\mathcal{D}\vert}\sum_{(x_i^k, y_i^k)\in\mathcal{D}}\sum_{m\in\mathcal{M}}\mathbf{1}_{(y_i^k=m)}t_i^k\cdot\\
& \log\tau_{m}(\mathbf{f}(x_i^k)),
\end{split}
\end{equation*}
where $t_i^k$ serves as the weight of each training instance. $\tau_{m}(\mathbf{f}(x_i^k))$ is the predicted probability of $x_i^k$ having the class label $m\in\mathcal{M}$, with the $d$-dimensional ``[CLS]'' token embeddings of the last layer of BERT (denoted as $\mathbf{f}(x_i^k)$) as features.

\subsection{Learning Domain-invariant Representations}


Based on previous research of domain-invariant learning~\cite{DBLP:conf/aaai/ShenQZY18,DBLP:conf/emnlp/HuWZGCS19}, we could add an additional domain adversarial classifier for MFT to optimize. However, we observe that adding such classifiers to models such as BERT may be sub-optimal. For ease of understanding, we only consider two domains $k_1$ and $k_2$. The loss of the adversarial domain classifier $\mathcal{L}_{AD}$ is:
\begin{equation*}
\begin{split}
\mathcal{L}_{AD}=&-\frac{1}{N^{k_1}+N^{k_2}}\sum_{(x_i^k,y_i^k)\in\mathcal{D}^{k_1}\cup\mathcal{D}^{k_2}}\\
&(y_i^k\log\sigma(x_i^k)+(1-y_i^k)\log(1-\sigma(x_i^k))),
\end{split}
\end{equation*}
where $y_i^k=1$ if the domain is $k_1$ and 0 otherwise. $\sigma(x_i^k)$ is the predicated probability of such adversarial domain classifier. In the min-max game of adversarial learning~\cite{DBLP:conf/aaai/ShenQZY18}, we need to maximize the loss $\mathcal{L}_{AD}$ such that the domain classifier fails to predict the true domain label. The min-max game between encoders and adversarial classifiers is computationally expensive, which is less appealing to MFT over large language models. Additionally, models such as BERT do not have the ``shared-private'' architecture~\cite{DBLP:conf/acl/LiuQH17}, frequently used for transfer learning.
One can also replace $\mathcal{L}_{AD}$ by asking the classifier to predict the flipped domain labels directly~\cite{DBLP:conf/iclr/ShuBNE18,DBLP:conf/emnlp/HuWZGCS19}. Hence, we can instead minimize the flipped domain loss $\mathcal{L}_{FD}$:
\begin{equation*}
\begin{split}
\mathcal{L}_{FD}=&-\frac{1}{N^{k_1}+N^{k_2}}\sum_{(x_i^k,y_i^k)\in\mathcal{D}^{k_1}\cup\mathcal{D}^{k_2}}\\
&((1-y_i^k)\log\sigma(x_i^k)+y_i^k\log(1-\sigma(x_i^k))).
\end{split}
\end{equation*}

We claim that, applying $\mathcal{L}_{FD}$ to BERT as an auxiliary loss does not necessarily generate domain-invariant features. When $\mathcal{L}_{FD}$ is minimized, $\sigma(x_i^k)$ always tends to predict the~\emph{wrong} domain label (which predicts $k_1$ for $k_2$ and $k_2$ for $k_1$). The optimization of $\mathcal{L}_{FD}$ still makes the learned features to be \emph{domain-dependent}, since the domain information is encoded implicitly in $\sigma(x_i^k)$, only with domain labels inter-changed.
A similar case holds for multiple domains where we only force the classifier to predict the domain of the input text $x_i^{k_j}$ into any one of the reminder $K-1$ domains (excluding $k_j$). 
Therefore, it is necessary to modify $\mathcal{L}_{FD}$ which truly guarantees domain invariance and avoids the expensive (and sometimes unstable) computation of adversarial training.

In this work, we propose the~\emph{domain corruption} strategy to address this issue. Given a training instance $(x_i^k, y_i^k)$ of the $k$th domain, we generate a corrupt domain label $z_i$ from a corrupted domain distribution $\Pr(z_i)$. $z_i$ is unrelated to the true domain label $k$, which may or may not be the same as $k$. The goal of the domain classifier is to approximate $\Pr(z_i)$ instead of always predicting the incorrect domains as in~\citet{DBLP:conf/emnlp/HuWZGCS19}. In practice, $\Pr(z_i)$ can be defined with each domain uniformly distributed, if the $K$ domain datasets are relatively balanced in size. To incorporate prior knowledge of domain distributions into the model, $\Pr(z_i)$ can also be non-uniform, with domain probabilities estimated from $\mathcal{D}$ by maximum likelihood estimation.

Consider the neural architecture of transformer encoders in BERT~\cite{DBLP:conf/naacl/DevlinCLT19}. Let $\mathbf{h}_l(x_i^k)$ be the $d$-dimensional mean pooling of the token embeddings of $x_i^k$ in the $l$th layer (excluding the ``[CLS]'' embeddings), i.e.,
\begin{equation*}
\mathbf{h}_l(x_i^k)=\text{Avg}(\mathbf{h}_{l,1}(x_i^k),\cdots,\mathbf{h}_{l,Max}(x_i^k)),
\end{equation*}
where $\mathbf{h}_{l,j}(x_i^k)$ represents the $l$-the layer embedding of the $j$th token in $x_i^k$, and $Max$ is the maximum sequence length. Additionally, we learn a $d$-dimensional domain embedding of the true domain label of $(x_i^k, y_i^k)$, denoted as $\mathcal{E}_D(k)$. The input features are constructed by adding the two embeddings: $\mathbf{h}_{l}(x_i^k)+\mathcal{E}_D(k)$, with the~\emph{typicality-sensitive domain corruption loss} $\mathcal{L}_{TDC}$ as:
\begin{equation*}
\begin{split}
 \mathcal{L}_{TDC}= &-\frac{1}{\vert\mathcal{D}\vert}\sum_{(x_i^k, y_i^k)\in\mathcal{D}}\sum_{k=1}^K\mathbf{1}_{(k=z_i)}t_i^k\\
& \cdot\log\delta_{z_i}{(\mathbf{h}_l(x_i^k)+\mathcal{E}_D(k))},
\end{split}
\end{equation*}
where $\delta_{z_i}(\cdot)$ is the predicted probability of the input features having the corrupted domain label $z_i$. We deliberately feed the true domain embedding $\mathcal{E}_D(k)$ into the classifier to make sure even if the classifier knows the true domain information from $\mathcal{E}_D(k)$, it can only generate corrupted outputs. In this way, we force the BERT's representations $\mathbf{h}_l(x_i^k)$ to hide any domain information from being revealed, making the representations of $x_i^k$ domain-invariant.

We further notice that neural language models may have deep layers. To improve the level of domain invariance of all layers and create a balance between effectiveness and efficiency, we follow the work~\cite{DBLP:conf/emnlp/SunCGL19} to train a series of~\emph{skip-layer} classifiers. Denote $L_{s}$ as the collection of selected indices of layers (for example, one can set $L_{s}=\{4, 8, 12\}$ for BERT-base). The~\emph{skip-layer domain corruption loss} $\mathcal{L}_{SDC}$ is defined as the average cross-entropy loss of all $\vert L_{s}\vert$ classifiers, defined as follows:
\begin{equation*}
\begin{split}
 \mathcal{L}_{SDC}= &-\frac{1}{\vert L_{s}\vert\cdot\vert\mathcal{D}\vert}\sum_{(x_i^k, y_i^k)\in\mathcal{D}}\sum_{l\in L_{s}}\sum_{k=1}^K\\
& \mathbf{1}_{(k=z_i)}t_i^k\cdot\log\delta_{z_i}{(\mathbf{h}_l(x_i^k)+\mathcal{E}_D(k))}.
\end{split}
\end{equation*}

In summary, the overall loss $\mathcal{L}$ for MFT to minimize is a linear combination of $\mathcal{L}_{TLC}$ and $\mathcal{L}_{SDC}$, i.e., $\mathcal{L}= \mathcal{L}_{TLC}+\lambda\mathcal{L}_{SDC}$, where $\lambda>0$ is a tuned hyper-parameter.

\begin{algorithm}[t]
\caption{Learning Algorithm for MFT}
\label{alg:meta}
\begin{small}
\begin{algorithmic}[1]
\STATE Restore BERT's parameters from the pre-trained model, with others randomly initialized;
\FOR {each domain $k$ and each class $m$}
\STATE Compute prototype embeddings $\mathbf{c}_m^k$;
\ENDFOR
\FOR {each training instance $(x_i^k, y_i^k)\in\mathcal{D}$}
\STATE Compute typicality score $t_i^k$;
\ENDFOR
\WHILE{number of training steps do not reach a limit}
\STATE Sample a batch $\{(x_i^k, y_i^k)\}$ from $\mathcal{D}$;
\STATE Shuffle domain labels of $\{(x_i^k, y_i^k)\}$ to generate $\{z_i\}$;
\STATE Estimate model predictions of inputs $\{(x_i^k, k)\}$ and compare them against $\{(y_i^k, z_i)\}$;
\STATE Update all model parameters by back propagation;
\ENDWHILE
\end{algorithmic}
\end{small}
\end{algorithm}

\subsection{Joint Optimization}

We describe how to optimize $\mathcal{L}$ for MFT. Based on the formation of $\mathcal{L}$, it is trivial to derive that:
\begin{equation*}
\begin{split}
\mathcal{L}= & -\frac{1}{\vert\mathcal{D}\vert}\sum_{(x_i^k, y_i^k)\in\mathcal{D}}(\sum_{m\in\mathcal{M}}\mathbf{1}_{(y_i^k=m)}t_i^k\cdot\\
& \log\tau_{m}(\mathbf{f}(x_i^k))+\frac{\lambda}{\vert L_{s}\vert}\sum_{l\in L_{s}}\sum_{k=1}^K\mathbf{1}_{(k=z_i)}t_i^k\cdot\\
&\log\delta_{z_i}{(\mathbf{h}_l(x_i^k)+\mathcal{E}_D(k))}).
\end{split}
\end{equation*}

As seen, MFT can be efficiently optimized via gradient-based algorithms with slight modifications. The procedure is shown in Algorithm~\ref{alg:meta}. It linearly scans the multi-domain training set $\mathcal{D}$ to compute prototypes $\mathbf{c}_m^k$ and typicality scores $t_i^k$. Next, it updates model parameters by batch-wise training. For each batch $\{(x_i^k, y_i^k)\}$, as an efficient implementation, we shuffle the domain labels to generate the corrupted labels $\{z_i\}$, as an approximation of sampling from the original corrupted domain distribution $\Pr(z_i)$. This trick avoids the computation over the whole dataset, and be adapted to the changes of domain distributions if new training data is continuously added to $\mathcal{D}$ through time. When the iterations stop, we remove all the additional layers that we have added for MFT, and fine-tune BERT for the $K$ domains over their respective training sets, separately.

\section{Experiments}

We conduct extensive experiments to evaluate MFT on multiple multi-domain text mining tasks.

\subsection{Datasets and Experimental Settings}

We employ the Google's pre-trained BERT model\footnote{We use the uncased, base version of BERT. See:~\url{https://github.com/google-research/bert}.} as the language model, with dimension $d=768$. Three multi-domain NLP tasks are used for evaluation, with statistics of datasets reported in Table~\ref{tab:dataset}:
\begin{itemize}
\item~\textbf{Natural language inference}: predicting the relation between two sentences as ``entailment'', ``neutral'' or ``contradiction'', using the dataset~\emph{MNLI}~\cite{DBLP:conf/naacl/WilliamsNB18}. MNLI is a large-scale benchmark corpus for evaluating natural language inference models, with multi-domain data divided into five genres.
\item~\textbf{Review analysis}: classifying the product review sentiment of the famous Amazon Review Dataset~\cite{DBLP:conf/acl/BlitzerDP07} (containing product reviews of four domains crawled from the Amazon website) as positive or negative.
\item~\textbf{Domain taxonomy construction}: predicting whether there exists a hypernymy (``is-a'') relation between two terms (words/noun phrases) for taxonomy derivation. Labeled term pairs sampled from three domain taxonomies are used for evaluation. The domain taxonomies are constructed by~\citet{DBLP:journals/coling/VelardiFN13}. with labeled datasets created and released by~\citet{DBLP:conf/emnlp/LuuTHN16}~\footnote{Following~\citet{DBLP:conf/emnlp/LuuTHN16}, in this task, we only do the binary classification of domain term pairs as hypernymy or non-hypernymy and do not consider reconstructing the graph structures of the domain taxonomies.} .
\end{itemize}

\begin{table}  
\centering
\begin{small} 
\begin{tabular}{lllll}  
\hline
\bf Dataset & \bf Domain &  \bf \#Train & \bf \#Dev & \bf \#Test\\
\hline
\multirow{5}{*}{MNLI} & Telephone & 83,348 & 2,000 & -\\
& Government & 77,350 & 2,000 & -\\
& Slate & 77,306 & 2,000 & -\\
& Travel & 77,350 & 2,000 & -\\
& Fiction & 77,348 & 2,000 & -\\
\hline
\multirow{4}{*}{Amazon} & Book & 1,763 & 120 & 117\\
& DVD & 1,752 & 120 & 128\\
& Electronics & 1,729 & 144 & 127\\
& Kitchen & 1,756 & 119 & 125\\
\hline
\multirow{3}{*}{Taxonomy} & Animal & 8,650 & 1,081 & 1,076\\
& Plant & 6,188 & 769 & 781\\
& Vehicle & 842 & 114  & 103\\
\hline
\end{tabular} 
\end{small}
\caption{Statistical summarization of datasets.} \label{tab:dataset} 
\end{table}  

Because MNLI  does not contain public labeled test sets that can be used for single-genre evaluation, we randomly sample 10\% of the training data for parameter tuning and report the performance of the original development sets. We hold out 2,000 labeled reviews from the Amazon dataset~\cite{DBLP:conf/acl/BlitzerDP07} and split them into development and test sets. As for the taxonomy construction task, because BERT does not naturally support word-pair classification, we combine a term pair to form a sequence of tokens separated by the special token ``[SEP]'' as input. The ratios of training, development and testing sets of the three domain taxonomy datasets are set as 80\%:10\%:10\%.

The default hyper-parameter settings of MFT are as follows: $\alpha=0.5$, $L_s=\{4,8,12\}$ and $\lambda=0.1$. During model training, we run $1\sim2$ epochs of MFT and further fine-tune the model in $2\sim4$ epochs for each domain, separately. The initial learning rate is set as $2\times10^{-5}$ in all experiments. 
The regularization hyper-parameters, the optimizer and the reminder settings are the same as in~\citet{DBLP:conf/naacl/DevlinCLT19}. In MFT, only 7K$\sim$11.5K additional parameters need to be added (depending on the number of domains), compared to the original BERT model. 
All the algorithms are implemented with TensorFlow and trained with NVIDIA Tesla P100 GPU. The training time takes less than one hour. For evaluation, we use~\emph{Accuracy} as the evaluation metric for all models trained via MFT and fine-tuning. All the experiments are conducted three times, with the average scores reported.

\begin{table*}  
\centering
\begin{small} 
\begin{tabular}{l | llllll}  
\hline
\bf Method &  \bf Telephone &  \bf Government & \bf Slate  & \bf Travel & \bf Fiction & \bf Average\\
\hline
BERT (S) & 82.5 & 84.9 & 78.2 & 83.1 & 82.0 & 82.1\\
BERT (Mix) & 83.1 & 85.2 & 79.3 & 85.1 & 82.4 & 83.0\\
BERT (MTL) & 83.8 & 86.1 & \bf 80.2 & 85.2 & 83.6 & 83.8\\
BERT (Adv) & 81.9 & 84.2 & 79.8 & 82.0 & 82.2 & 82.0\\		
\hline			
MFT (DC) & \bf 84.2 & \bf 86.3 & \bf 80.2 & \bf 85.8 & \bf 84.0 & \bf 84.1\\
MFT (TW) & 83.8 & \bf 86.5 & \bf 81.3 & 83.7 & \bf 84.4 & \bf 83.9\\
MFT (Full) & \bf 84.6 & \bf 86.3 & \bf 81.5 & \bf 85.4 & \bf 84.6 & \bf 84.5\\
\hline
\end{tabular}
\end{small} 
\caption{Natural language inference results over MNLI (divided into five genres) in terms of accuracy (\%).} \label{tab:mnli} 
\end{table*}

\subsection{General Experimental Results}

We report the general testing results of MFT. For fair comparison, we implement following the fine-tuning methods as strong baselines:
\begin{itemize}
\item~\textbf{BERT (S)}: It fine-tunes $K$ BERT models, each with single-domain data.
\item~\textbf{BERT (Mix)}: It combines all the domain data and fine-tunes a single BERT model only.
\item~\textbf{BERT (MTL)}: It fine-tunes BERT by multi-task fine-tuning~\cite{DBLP:conf/cncl/SunQXH19}.
\item~\textbf{BERT (Adv)}: It fine-tunes BERT by BERT (C) with an additional adversarial domain loss proposed in~\citet{DBLP:conf/emnlp/HuWZGCS19}.
\end{itemize}

We also evaluate the performance of MFT and its variants under the following three settings:
\begin{itemize}
\item~\textbf{MFT (DC)}: It is MFT with domain corruption. All the typicality scores in the objective function are removed. 
\item~\textbf{MFT (TW)}: It is MFT with typicality weighting. The domain corruption loss is removed. 
\item~\textbf{MFT (Full)}: It is the full implementation.
\end{itemize}

\begin{table}  
\centering
\begin{small} 
\begin{tabular}{l | lllll}  
\hline
\bf Method & \bf Book &  \bf DVD & \bf Elec. & \bf Kit. & \bf Avg.\\
\hline
BERT (S) & 90.7 & 88.2 & 89.0 & 85.7 & 88.4\\
BERT (Mix) & 91.8  & \bf 89.4 & 87.8  & 88.4 & 89.3\\		
BERT (MTL) & \bf 92.2 & 89.0 & 88.3 & 88.2 & 89.0\\
BERT (Adv) & 89.3 & 87.4 & 86.5 & 86.7 & 87.5\\			
\hline			
MFT (DC) & 90.6 & \bf 89.4 & \bf 92.5 & \bf 88.7 & \bf 90.3\\
MFT (TW) & 90.4 & 88.9 & \bf 94.5 & \bf 89.1 & \bf 90.7\\
MFT (Full) & 91.2 & 88.8 & \bf 94.8 & \bf 89.2 & \bf 91.0\\
\hline
\end{tabular}
\end{small} 
\caption{Review analysis results over Amazon Review Dataset in terms of accuracy (\%).} \label{tab:sent} 
\end{table}  

The results of three multi-domain NLP tasks are reported in Table~\ref{tab:mnli}, Table~\ref{tab:sent} and Table~\ref{tab:tax}, respectively. Generally, the performance trends of all three tasks are pretty consistent. With MFT, the accuracy of fine-tuned BERT boosts 2.4\% for natural language inference, 2.6\% for review analysis and 3\% for domain taxonomy construction. The simple multi-task fine-tuning methods do not have large improvement, of which the conclusion is consistent with~\citet{DBLP:conf/cncl/SunQXH19}. Our method has the highest performance in 10 domains (genres) out of a total of 12 domains of the three tasks, outperforming previous fine-tuning approaches. For ablation study, we compare MFT (DC), MFT (TW) and MFT (Full). The results show that domain corruption is more effective than typicality weighting in MNLI, but less effective in Amazon and Taxonomy.

\begin{table}  
\centering
\begin{small} 
\begin{tabular}{l | llll}  
\hline
\bf Method & \bf Animal &  \bf Plant & \bf Vehicle & \bf Avg.\\
\hline
BERT (S) & 93.6 & \bf 91.8 & 84.2 & 89.3\\
BERT (Mix) & 93.8 & 88.2 & 83.6 & 88.5\\	
BERT (MTL) & 94.2 & 89.2 & 82.4 & 88.6\\	
BERT (Adv) & 92.8 & 86.3 & 83.2 & 87.4\\			
\hline			
MFT (DC) & \bf 94.3 & \bf 91.8 & \bf 86.8 & \bf 91.0\\
MFT (TW) & 94.0 & \bf 92.0 & \bf 89.2 & \bf 91.7\\
MFT (Full) & \bf 94.5 & \bf 92.3 & \bf 90.2 & \bf 92.3\\
\hline
\end{tabular}
\end{small} 
\caption{Domain taxonomy construction results over Taxonomy Dataset in terms of accuracy (\%).} \label{tab:tax} 
\end{table}

\begin{table*}  
\centering
\begin{small} 
\begin{tabular}{llll}  
\hline
\bf Typicality & \bf Domain & \bf Label & \bf Review Text\\
\hline
& Book & NEG & More hate books. How could anyone write anything so wrong.\\
Low & Kitchen & NEG &The spoon handle is crooked and there are marks/damage to the wood. Avoid.\\
& Kitchen & NEG &The glass is quite fragile. I had two breaks.\\
\hline
& Kitchen & POS &I would recommend them to everyone..and at their price, it's a HUGE DEAL! \\
High & Electronics & NEG & What a waste of money. For \$300 you shouldn't HAVE to buy a protection plan for... \\
& Electronics & NEG & Do not waste your money, this was under recommendations, but I would NOT... \\
\hline
\end{tabular}
\end{small} 
\caption{Cases of review texts from Amazon Review Dataset with high and low typicality scores.} \label{tab:case} 
\end{table*}  

\begin{table*}  
\centering
\begin{small} 
\begin{tabular}{l lll lll lll}  
\hline
\multirow{2}{*}{\bf Genre} & \multicolumn{2}{c}{\bf With MFT?} & \multirow{2}{*}{\bf Improvement} & \multicolumn{2}{c}{\bf With MFT?} & \multirow{2}{*}{\bf Improvement} & \multicolumn{2}{c}{\bf With MFT?} & \multirow{2}{*}{\bf Improvement}\\
\cline{2-3}\cline{5-6}\cline{8-9}
& \bf No & \bf Yes & &  \bf No & \bf Yes & & \bf No & \bf Yes\\
\hline
\bf Training data & \multicolumn{3}{c}{\textbf{5\%} of the original} & \multicolumn{3}{c}{\textbf{10\%} of the original} & \multicolumn{3}{c}{\textbf{20\%} of the original}\\
\hline
Telephone & 70.5 & 74.7 & \bf 4.2\%$\uparrow$ & 74.1 & 76.4 & \bf 2.3\%$\uparrow$ & 75.9 & 79.8 & \bf 3.9\%$\uparrow$\\
Government & 76.5 & 78.1 & \bf 1.6\%$\uparrow$ & 78.8 & 81.0 & \bf 2.2\%$\uparrow$ & 80.5 & 82.9 & \bf 2.4\%$\uparrow$\\
Slate & 64.2 & 69.8 & \bf 5.7\%$\uparrow$ & 67.6 & 71.8 & \bf 4.2\%$\uparrow$ & 71.8 & 74.1 & \bf 2.3\%$\uparrow$\\
Travel & 71.9 & 75.4 & \bf 3.5\%$\uparrow$ & 74.8 & 78.1 & \bf 3.3\%$\uparrow$ & 78.3 & 80.3 & \bf 2.0\%$\uparrow$\\
Fiction & 69.7 & 73.8 & \bf 4.1\%$\uparrow$ & 73.3 & 76.6 & \bf 3.3\%$\uparrow$ & 76.2 & 78.4 & \bf 2.2\%$\uparrow$\\
\hline
\bf Average & 70.5 & 74.4 & \bf 3.9\%$\uparrow$ & 73.7 & 76.8 & \bf 3.1\%$\uparrow$ & 76.5 & 79.1 & \bf 2.6\%$\uparrow$\\
\hline
\end{tabular}
\end{small} 
\caption{Few-shot natural language inference results over MNLI in terms of accuracy (\%).} \label{tab:few} 
\end{table*}  

\subsection{Detailed Model Analysis}

\begin{figure}
\centering
\subfigure[Dataset: Amazon]{
\includegraphics[width=0.35\textwidth]{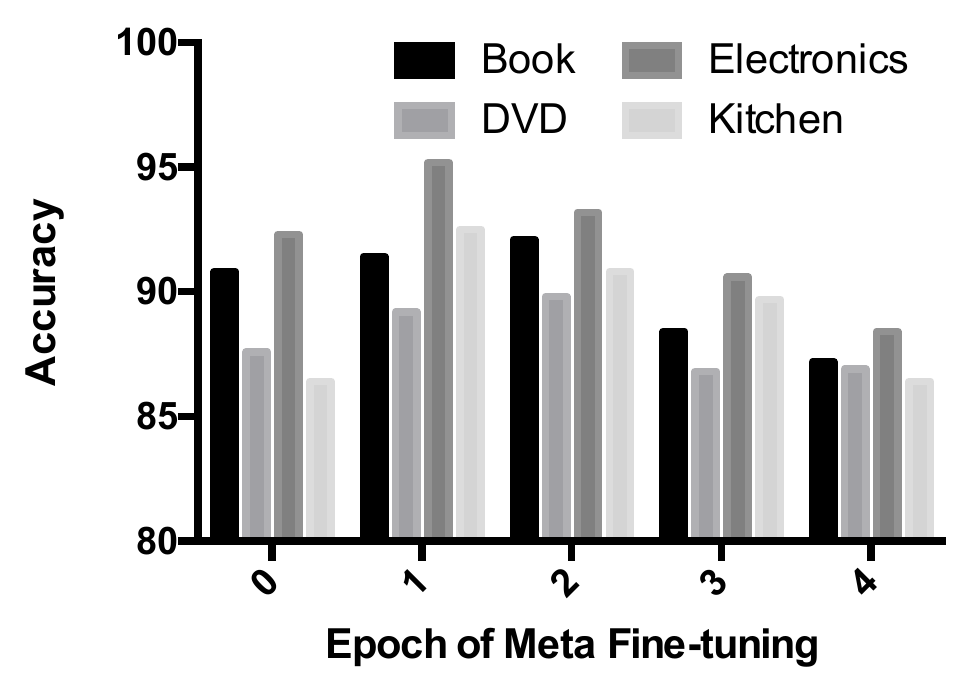}}
\subfigure[Dataset: Taxonomy]{
\includegraphics[width=0.35\textwidth]{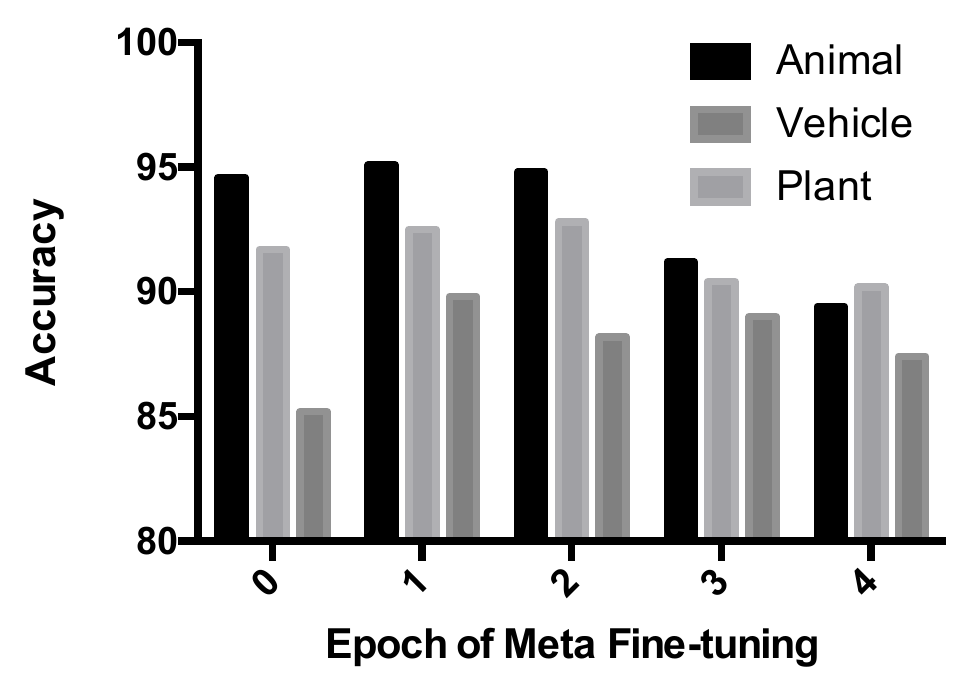}}
\caption{Tuning the learning epochs of MFT.} \label{fig:epoch} 
\end{figure}

\begin{figure}
\centering
\subfigure[Dataset: Amazon]{
\includegraphics[width=0.35\textwidth]{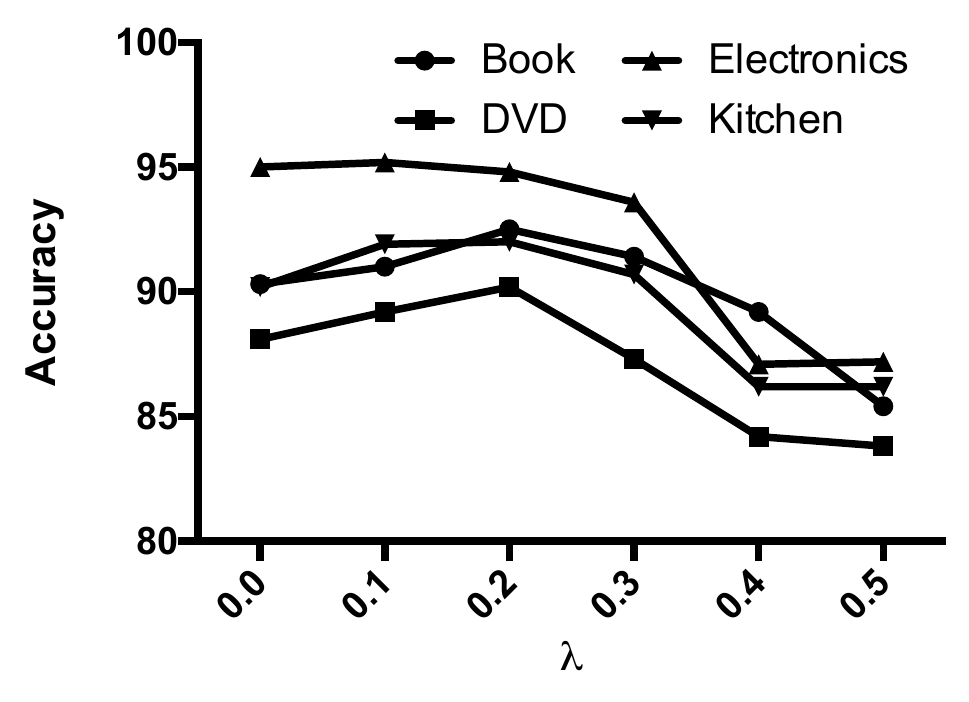}}
\subfigure[Dataset: Taxonomy]{
\includegraphics[width=0.35\textwidth]{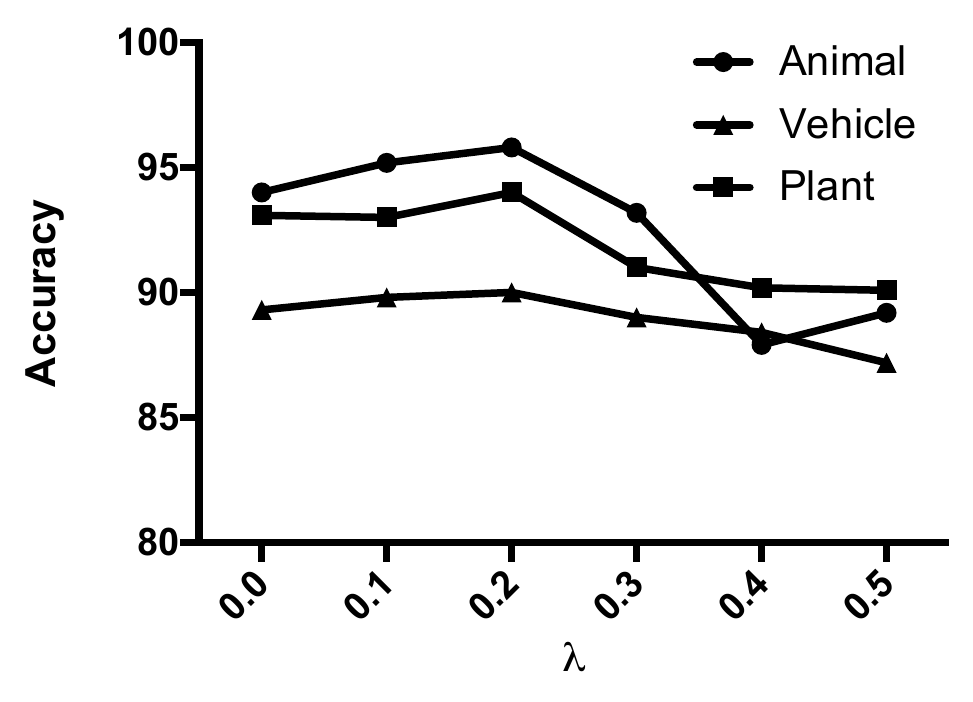}}
\caption{Tuning the hyper-parameter $\lambda$.} \label{fig:lam} 
\end{figure}

In this section, we present more experiments on detailed analysis of MFT. We first study how many training steps of MFT we should do before fine-tuning. As datasets of different tasks vary in size, we tune the epochs of MFT instead. In this set of experiments, we fix parameters as default, vary the training epochs of MFT and then run fine-tuning for 2 epochs for all domains. The results of two NLP tasks are shown in Figure~\ref{fig:epoch}. It can be seen that too many epochs of MFT can hurt the performance because BERT may learn too much from other domains before fine-tuning on the target domain. We suggest that a small number of MFT epochs are sufficient for most cases. Next, we tune the hyper-parameter $\lambda$ from 0 to 0.5, with the results shown in Figure~\ref{fig:lam}. The inverted-V trends clearly reflect the balance between the two types of losses in MFT, with very few exceptions due to the fluctuation of the stochastic learning process.

We also vary the number of corrupted domain classifiers by changing $L_{s}$. Due to space limitation, we only report averaged accuracy across all domains, shown in Table~\ref{tab:layer}. In a majority of scenarios, adding more corrupted domain classifiers slightly improve the performance, as it poses strong domain-invariance constraints to deeper layers of transformer encoders in BERT.

\begin{table}  
\centering
\begin{small} 
\begin{tabular}{l | ll}  
\hline
\bf $L_{s}\downarrow$ Dataset$\rightarrow$ & \bf Amazon & \bf Taxonomy\\
\hline
$\{12\}$ & 90.7 & 91.3\\
$\{6,12\}$ & 90.8 & 92.0\\
$\{4,8,12\}$ & 91.4 & 92.5\\
$\{3,6,9,12\}$ & 91.2 & 92.8\\
\hline			
\end{tabular}
\end{small} 
\caption{Change of prediction accuracy when the selected layer indices $L_{s}$ take different values (\%).} \label{tab:layer} 
\end{table}

For more intuitive understanding of MFT, we present some cases from Amazon Review Dataset with relatively extreme (low and high) typicality scores, shown in Table~\ref{tab:case}. As seen, review texts with low scores are usually related to certain aspects of specific products (for example, ``crooked spoon handle'' and ``fragile glass''), whose knowledge is non-transferable on how to do review analysis in general. In contrast, reviews with high typicality scores may contain expressions on the review polarity that can be frequently found in various domains (for example, ``huge deal'' and ``a waste of money''). From the cases, we can see how MFT can create a meta-learner that is capable of learning to solve NLP tasks in different domains.

\subsection{Experiments for Few-shot Learning}

Acquiring a sufficient amount of training data for emerging domains often poses challenges for NLP researchers. In this part, we study how MFT can benefit few-shot learning when the size of the training data in a specific domain is small~\footnote{Note that this experiment is not conducted using the K-way N-shot setting. Refer to Related Work for discussion.}. Because the original MNLI dataset~\cite{DBLP:conf/naacl/WilliamsNB18} is large in size, we randomly sample 5\%, 10\% and 20\% of the original training set for each genre, and do MFT and fine-tuning over BERT. For model evaluation, we use the entire development set without sampling. In this set of experiments, we do not tune any hyper-parameters and set them as default. Due to the small sizes of our few-shot training sets, we run MFT for only one epoch, and then fine-tune BERT for 2 epochs per genre.

In Table~\ref{tab:few}, we report the few-shot learning results, and compare them against fine-tuned BERT without MFT. From the experimental results, we can safely come to the following conclusions. i) MFT improves the performance for text mining of all genres in MNLI, regardless of the percentages of the original training sets we use. ii) MFT has a larger impact on smaller training sets (a 3.9\% improvement in accuracy for 5\% few-shot learning, compared to a 2.6\% improvement for 20\%). iii) The improvement of applying MFT before fine-tuning is almost the same as doubling the training data size.  Therefore, MFT is highly useful for few-shot learning when the training data of other domains are available.

\section{Conclusion and Future Work}

In this paper, we propose a new training procedure named Meta Fine-Tuning (MFT) used in neural language models for multi-domain text mining. Experimental results show the effectiveness of MFT from various aspects. In the future, we plan to apply MFT to other language models (e.g.,~Transformer-XL~\cite{DBLP:conf/acl/DaiYYCLS19} and ALBERT~\cite{DBLP:journals/corr/abs-1909-11942}) and for other NLP tasks.

\section*{Acknowledgements}
We would like to thank anonymous reviewers for
their valuable comments. 
This work is partially supported by
the National Key Research and Development Program of China under Grant No. 2016YFB1000904.
M. Qiu is partially funded by China Postdoctoral Science Foundation (No. 2019M652038).

\end{document}